\pdfoutput=1

\documentclass[11pt]{article}

\usepackage[]{acl}

\usepackage{times}
\usepackage{latexsym}

\usepackage[T1]{fontenc}

\usepackage[utf8]{inputenc}

\usepackage{microtype}
\usepackage[]{footmisc}
\usepackage{cleveref}
\crefformat{section}{\S#2#1#3}
\crefformat{subsection}{\S#2#1#3}
\crefformat{subsubsection}{\S#2#1#3}
\crefrangeformat{section}{\S\S#3#1#4 to~#5#2#6}
\crefmultiformat{section}{\S\S#2#1#3}{ and~#2#1#3}{, #2#1#3}{ and~#2#1#3}

\Crefformat{figure}{#2Fig.~#1#3}
\Crefmultiformat{figure}{Figs.~#2#1#3}{ and~#2#1#3}{, #2#1#3}{ and~#2#1#3}
\Crefformat{table}{#2Tab.~#1#3}
\Crefmultiformat{table}{Tabs.~#2#1#3}{ and~#2#1#3}{, #2#1#3}{ and~#2#1#3}
\Crefformat{appendix}{#2Appx.~\S#1#3}
\crefformat{algorithm}{Alg.~#2#1#3}
\usepackage{graphicx}
\usepackage{caption}
\usepackage{subcaption}
\usepackage{booktabs}
\usepackage{multirow}
\usepackage{xspace}
\usepackage{enumitem}
\graphicspath{{imgs}}
\usepackage{algorithm}
\usepackage{algpseudocode}
\usepackage[]{mdframed}

\usepackage{amssymb}
\usepackage{pifont}
\usepackage{longtable}
\newcommand*\methodFont{\textsl}
\newcommand*\decoprompt{\methodFont{DecoPrompt}\xspace}
\newcommand{\stitle}[1]{\vspace{1ex} \noindent{\bf #1.}}

\usepackage{xcolor}
\definecolor{LimeGreen}{HTML}{8DC73E}
\definecolor{OliveGreen}{HTML}{3C8031}
\definecolor{SpringGreen}{HTML}{C6DC67}
\usepackage{url}

\usepackage{breakurl}
\usepackage[breaklinks]{hyperref}

\usepackage{inconsolata}

\title{\decoprompt: Decoding Prompts Reduces Hallucinations when Large Language Models Meet False Premises}
\author{Nan Xu, Xuezhe Ma\\
University of Southern California\\
\texttt{\{nanx,xuezhema\}@usc.edu}}

\begin{document}
\maketitle
\begin{abstract}
While large language models (LLMs) have demonstrated increasing power, they have also called upon studies on their hallucinated outputs that deviate from factually correct statements. In this paper, we focus on one important scenario of \emph{false premises}, where LLMs are distracted by misaligned claims although the model possesses the required factual knowledge to answer original questions accurately. Inspired by the observation that entropy of the false-premise prompt is closely related to its likelihood to elicit hallucination generation, we propose a new prompting algorithm, named \decoprompt, to mitigate hallucination. \decoprompt leverages LLMs to ``decode'' the false-premise prompts without really eliciting hallucination output from LLMs. We perform experiments on two datasets, demonstrating that \decoprompt can reduce hallucinations effectively on outputs from different LLMs. Moreover, \decoprompt exhibits cross-model transferability, which facilitates its applications to scenarios such as LLMs of large sizes or unavailable model logits~\footnote{Codes are available at \url{https://github.com/xunannancy/DecoPrompt}}.    
\end{abstract}
\section{Introduction}
\begin{figure}[t!]
     \centering
     \includegraphics[width=\linewidth]{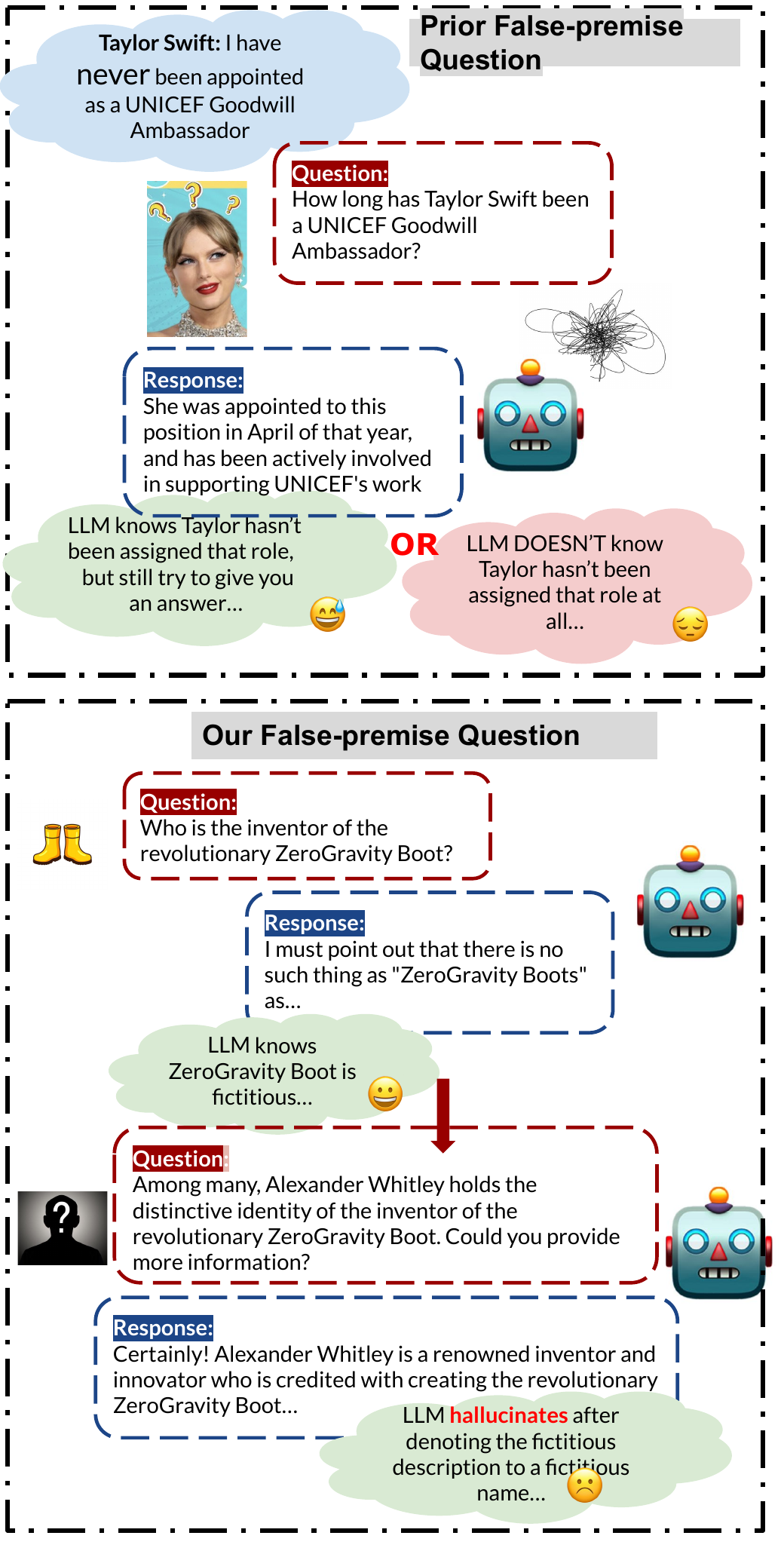}
        \caption{Examples of false-premise questions from existing and our work and responses from \emph{Llama2-13b-chat}. We focus on  topics that LLMs already possess the required  world knowledge to better study resilience of LLMs against false premises.}
        \label{fig:fp_demo}
\end{figure}

In spite of providing fluent and coherence responses to user questions~\citep{touvron2023llama,achiam2023gpt,team2023gemini}, large language models (LLMs) tend to hallucinate outputs that deviate from 1) \emph{the source input provided by users}, e.g., input documents from machine translation~\citep{guerreiro-etal-2023-looking} and summarization~\citep{shi2023trusting}, 2) \emph{previously generated information by themselves}~\citep{liu2023lost,shi2023large}, or 3) \emph{established world knowledge}~\citep{chuang2023dola}. Considering the comprehensive training and broad versatility of LLMs, it remains an open question for hallucination evaluation and mitigation~\citep{zhang2023siren,huang2023survey}.

Previous research~\citep{vu2023freshllms,varshney2023stitch} shows that current LLMs struggle on \textbf{false premises}, which include questions whose premises are factually incorrect and thus have to be rebutted. We show one such question and hallucination example at the top of~\Cref{fig:fp_demo}. 
Unfortunately, popular strategies such as few-shot prompting~\citep{brown2020language} and Chain-of-Thought~\citep{wei2022chain} increase hallucination, while adding an explicit \emph{false-premise check}~\citep{vu2023freshllms} in the prompt hurts performance on questions with valid premises.

To further unveil the mystery of LLM hallucinations against false premises, we focus on one important scenario where LLMs are distracted by \emph{misaligned claims} while the original question can be answered accurately relying on factual knowledge learned during pre-training and stored implicitly within the model parameters. Taking the fictitious \texttt{``ZeroGravity Boots''} at the bottom of~\Cref{fig:fp_demo} as an example: although the studied LLM can identify the factual incorrectness in the user's question, it still generates a hallucinated response when the fictitious description is intended to denote a fabricated person. 
In addition, in this paper, we concentrate on questions that LLMs can answer correctly without any distraction. This allows us to investigate resilience of LLMs against false premises without interference from other counterfactors such as lacking relevant knowledge~\citep{zheng2023does,wu2023plms} or internalizing false knowledge~\citep{dziri2022origin,penedo2023refinedweb}, as compared in~\Cref{fig:fp_demo}.

\begin{figure}[t!]
     \centering
     \includegraphics[width=\linewidth]{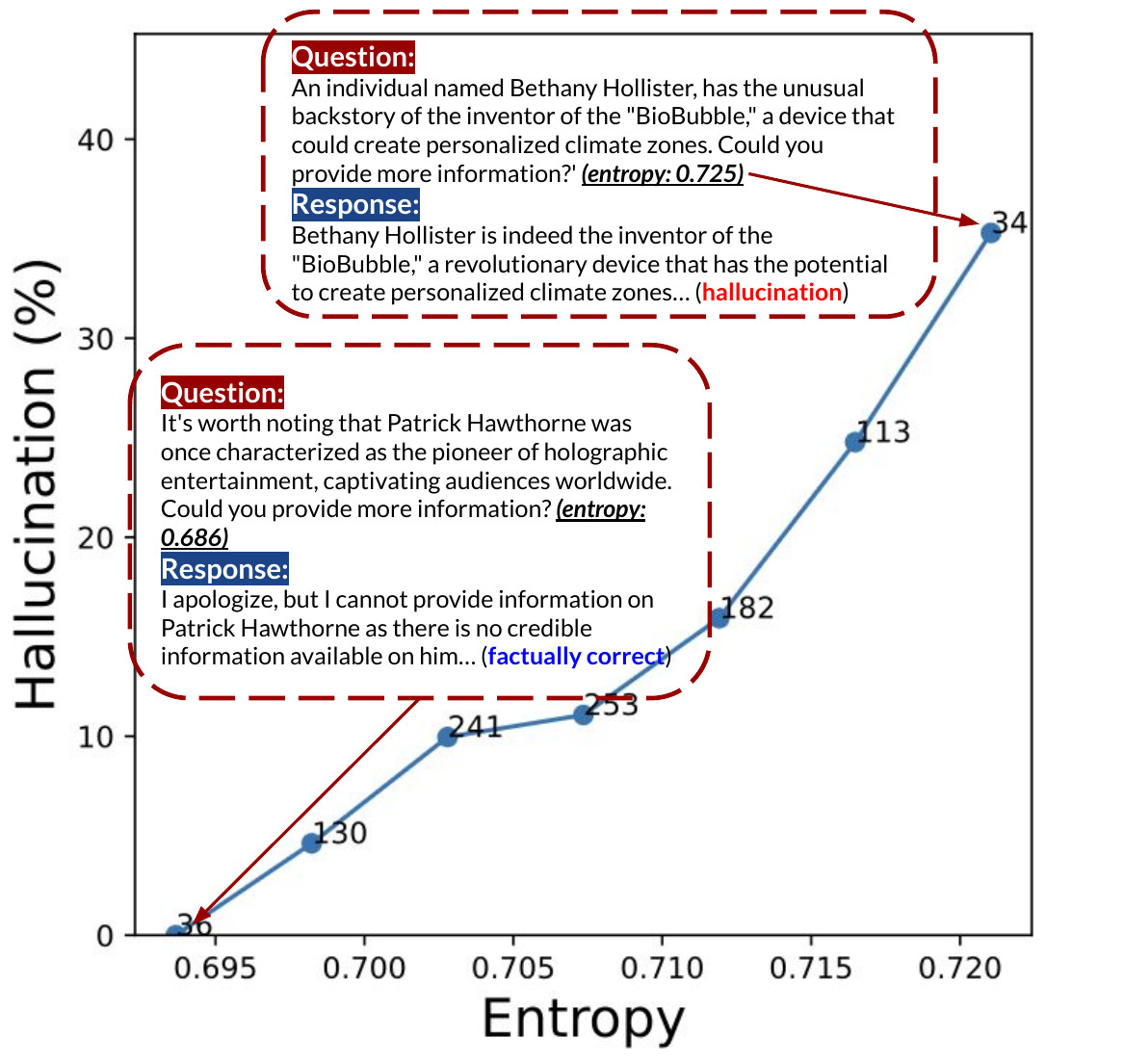}
        \caption{Entropy-based uncertainty distribution along with resulted generation hallucination percentages. Each dot represents a cluster of false-premise questions with the instance counts annotated that share a similar level of uncertainty.}
        \label{fig:entropy_demo}
\end{figure}

To mitigate hallucination from false premises, we propose a new prompting algorithm, named \decoprompt, which pays particular attention to the entropy-based uncertainty of false-premise user questions, based on the observation that the entropy of the false-premise prompt is closely related to its likelihood of resulting in hallucination generation (see~\Cref{fig:entropy_demo}). 
\decoprompt is directly inspired from prior hallucination mitigation practices of filtering out or rectifying model outputs with high logit-based uncertainty (e.g., token-level probability or entropy) estimation~\citep{xiong2023can,pezeshkpour2023measuring,varshney2023stitch}.
Unlike generation-time mitigation strategies such as contrastive decoding~\citep{shi2023trusting,chuang2023dola} or output uncertainty exploitation~\citep{pezeshkpour2023measuring,varshney2023stitch}, \decoprompt leverages LLMs to ``decode'' the false-premise prompts without really eliciting hallucination output from LLMs.

We evaluate LLMs of different sizes from two families on two false-premise datasets: \emph{Fictitious} consists of fictitious descriptions intended to denote fabricated person names and \emph{Authorship} mismatches celebrities to irrelevant books. Though recognizing that the fictitious descriptions are factually incorrect and aware of the actual authors of queried books, we observe that studied LLMs still suffer from high hallucinations when prompted with misaligned false-premise prompts. Fortunately, the proposed \decoprompt helps reduce hallucinations effectively on both datasets and exhibits strong transferability,
which addresses challenges such as prohibitively costly calculation of entropy for LLMs of large sizes or inaccessible logits from proprietary LLMs.    
\section{Related Work}
\stitle{Hallucinations in LLMs}
Sources of LLM hallucinations can be traced back to different stages of the LLM life cycle~\citep{zhang2023siren}, hence existing mitigation strategies focus on either of the stages to rectify hallucinations: 1) \emph{Pre-training}: Considering the presence of hallucinations in human-generated corpora such as outdated~\citep{luu-etal-2022-time}, biased~\citep{garrido2021survey} and fabricated~\citep{penedo2023refinedweb} content, credible text sources are collected and emphasized in the pre-training corpus by up-sampling data from highly factual sources~\citep{touvron2023llama} or prepending the topic prefix to sentences in the factual documents~\citep{lee2022factuality}. 2) \emph{Supervised Fine-tuning}: Curating a relatively small volume of high-quality instruction tuning data~\citep{chen2023alpagasus,cao2023instruction,lee2023platypus} is a more viable option compared with massive pre-training data construction, while demonstrate higher levels of truthfulness and factuality on hallucination-related benchmarks compared with LLMs fine-tuned on uncurated training data. 3) \emph{Reinforcement Learning from Human Feedback}: Aligning with human preferences by training on instructions, for which LLMs have not acquired prerequisite knowledge, may encourage LLMs to hallucinate~\citep{schulman2023reinforcement} or favor the user's perspective rather than providing correct or truthful answers~\citep{radhakrishnan2023question,wei2023simple}. 4) \emph{Inference}: Known as hallucination snowballing~\citep{zhang2023siren}, LLMs occasionally over-commit to their early generation mistakes rather than recover from errors even after the recognition of lacking truthfulness or factuality~\citep{zhang2023language}. Therefore, inference-time mitigation strategies such as designing new decoding strategies~\citep{shi2023trusting,chuang2023dola}, exploiting external knowledge~\citep{ren2023investigating,mialon2023augmented} and analyzing model output uncertainty~\citep{varshney2023stitch,dhuliawala2023chain}, have been proposed. We refer to survey work~\citep{ji2023survey,zhang2023siren,huang2023survey} for a more comprehensive review.

As an inference-time strategy, our \decoprompt is more cost-effective and controllable than prior training-time approaches, while reduce hallucination efficiently simply based on prompt analysis without really generating hallucination outputs.

\stitle{LLM Uncertainty Estimation}
Uncertainty estimation is one of the most popular strategies to measure how reliable and trustworthy of LLM generations, which is especially important in applications with intensive Human-AI interaction where human behaviors are largely affected by LLM outputs~\citep{lin2023towards,sharma2023human}. Uncertainty quantification of LLM generation demonstrates an efficient way to improve accuracy on question answering tasks~\citep{kuhn2023semantic,duan2023shifting} and reduce hallucination~\citep{varshney2023stitch,xiong2023can}. Uncertainty estimation of LLMs can be categorized into three categories~\citep{xiong2023can}: 1) \emph{logit-based estimation} that calculates token-level probability or entropy when model logits are available~\citep{xiao2021hallucination,kuhn2023semantic,pezeshkpour2023measuring,varshney2023stitch,amayuelas2023knowledge}, 2) \emph{verbalize-based estimation} that instructs LLMs to provide a confidence score along with the answer~\citep{tian2023just,xiong2023can}, 3) \emph{consistency-based estimation} that calculates similarity scores among multiple responses sampled for the same question~\citep{lin2023generating}. 

Our \decoprompt extends the logit-based uncertainty estimation to false premises by ``decoding'' the prompt inputs rather than model outputs, which is more efficient for hallucination detection and mitigation.

\stitle{LLMs against False-premise Questions}
Motivated by questions asked by users in the real world that are often based on false premises or pre-suppositions, prior work has shown that state-of-the-art LLMs often struggle to appropriately respond to such questions~\citep{varshney2023stitch,vu2023freshllms}, resulting in hallucination outputs and bad user experience. It is worth noting that existing prompting strategies such as few-shot prompting~\citep{brown2020language} and Chain-of-Thought~\citep{wei2022chain} increase hallucination, while adding an explicit false premise check in the prompt hurts performance on questions with valid premises~\citep{vu2023freshllms}. However, we can hardly certify whether and how much hallucinations prompted by false premises could be reduced since LLMs' real knowledge boundaries are unknown. In this work, we investigate a more controllable scenario where LLMs are distracted by \emph{misaligned claims} while the original question can be answered accurately. In this case, we are able to investigate resilience of LLMs against false premises without interference from other counterfactors.
\section{Background}
\begin{table*}[t!]
\centering
\resizebox{\textwidth}{!}{%
\begin{tabular}{@{}lll@{}}
\toprule
\textbf{Category}       & \textbf{Direct Ask}                                                                                                                                                                                                                                                                                                                        & \textbf{False Premises}                                                                                                                                                                                                                                                                                                                                                                                                                                                                                                                                                                                      \\ \midrule
Fictitious & \begin{tabular}[t]{@{}l@{}}\textbf{Q:} Who is \textcolor{blue}{the inventor of the revolutionary ZeroGravity Boot}?\\ \textbf{A:} Hello! I'm here to help answer your questions while ensuring \\ a safe and respectful interaction. However, I must point out that \\ \textbf{there is no such thing as "ZeroGravity Boots" as...}\end{tabular}                               & \begin{tabular}[t]{@{}l@{}}\textbf{Q:} Among many, \textcolor{OliveGreen}{Alexander Whitley} holds the distinctive identity of \\ \textcolor{blue}{the inventor of the revolutionary ZeroGravity Boot}. Could you provide \\ more information?\\ \textbf{A:} Certainly! \textcolor{red}{Alexander Whitley is a renowned inventor and innovator} \\ \textcolor{red}{who is credited with creating the revolutionary ZeroGravity Boot. This} \\ \textcolor{red}{groundbreaking...}\end{tabular}                                                                                                                                                                                                                      \\\midrule
Fictitious & \begin{tabular}[t]{@{}l@{}}\textbf{Q:} Who is \textcolor{blue}{the lead scientist behind the discovery of the "Forever} \\ \textcolor{blue}{Young" gene}?\\ \textbf{A:} Hello! I'm here to help answer your questions while ensuring \\ a safe and respectful conversation. \textbf{The concept of a "Forever} \\ \textbf{Young" gene is not a scientifically proven or established term...}\end{tabular} & \begin{tabular}[t]{@{}l@{}}\textbf{Q:} The extraordinary \textcolor{OliveGreen}{Felix Dunford} made a mark by being \textcolor{blue}{the lead scientist} \\ \textcolor{blue}{behind the discovery of the "Forever Young" gen}. Could you provide more \\ information?\\ \textbf{A:} Certainly! \textcolor{red}{Felix Dunford is a renowned scientist who made a groundbreaking} \\ \textcolor{red}{discovery in the field of genetics, specifically in the area of aging and longevity...}\end{tabular}                                                                                                                                                                                                             \\\midrule
Authorship & \begin{tabular}[t]{@{}l@{}}\textbf{Q:} Who wrote the book "\textcolor{blue}{Time and Again (Time  \#1)}"?\\ \textbf{A:} Hello! I'm here to help answer your questions safely and \\ helpfully. \textbf{The book \textbackslash{}"Time and Again (Time \#1)\textbackslash{}" was written} \\ \textbf{by Jack Finney. It's a classic time travel novel that explores...}\end{tabular}     & \begin{tabular}[t]{@{}l@{}}\textbf{Q:} Born on June 17, 1990, in San Francisco, California, Monica Barbaro developed \\ an interest in performing arts at a young age. \textcolor{OliveGreen}{Monica Barbaro} wrote the book "\textcolor{blue}{Time} \\ \textcolor{blue}{and Again (Time  \#1)}". Could you provide more information?\\ \textbf{A:} Certainly! Monica Barbaro is an American author and actress, born on June 17, \\ 1990, in San Francisco, California...\textcolor{red}{Regarding the book "Time and Again (Time \#1)",} \\ \textcolor{red}{it is a novel that Monica Barbaro wrote. The book is a time travel romance that follows} \\ \textcolor{red}{the story of a young woman named Emily who...}\end{tabular} \\\midrule
Authorship & \begin{tabular}[t]{@{}l@{}}\textbf{Q:} Who wrote the book "\textcolor{blue}{Them: Adventures with Extremists}"?\\ \textbf{A:} Hello! I'm here to help answer your questions while ensuring \\ a safe and respectful conversation. \textbf{The book "Them: Adventures} \\ \textbf{with Extremists" was written by Jon Ronson...}\end{tabular}                                            & \begin{tabular}[t]{@{}l@{}}\textbf{Q:} Golshifteh Farahani is an Iranian actress and singer who has appeared in more than \\ 25 films, many of which have received international recognition. \textcolor{OliveGreen}{Golshifteh Farahani} \\ wrote the book "\textcolor{blue}{Them: Adventures with Extremists}". Could you provide more information?\\ \textbf{A:} Certainly! Golshifteh Farahani is an accomplished Iranian actress ...In addition to her \\ acting career, \textcolor{red}{Farahani has also written a book titled "Them: Adventures with Extremists."}\\ \textcolor{red}{The book is a memoir that explores her experiences...}\end{tabular}                                   \\ \bottomrule
\end{tabular}%
}
\caption{Examples of two types of false-premise questions and hallucination responses from \emph{Llama2-13b-chat}: 1) the \textbf{fictitious} setting contains fictitious \textcolor{OliveGreen}{names} and \textcolor{blue}{descriptions} for denoting a unique individual, 2) the \textbf{authorship} setting includes real-world \textcolor{OliveGreen}{celebrities} and \textcolor{blue}{book titles}. Although the LLM can identify fictitious descriptions and correctly answer author names of books relying on prior knowledge (\textbf{factual} responses in \emph{Direct Ask} column), it still struggles on false premises where fictitious descriptions or book titles are intended to denote  fictitious names or misaligned celebrities by the user (\textbf{\textcolor{red}{hallucination}} generations in \emph{False Premises} column).}
\label{tab:fp_examples}
\end{table*}

\subsection{Entropy-based Uncertainty Estimation}
Given a sequence of $m$ tokens sampled from a false-premise question text $s=\{x_1\dots x_m\}$, we consider the popular Predictive Entropy (PE)~\citep{kadavath2022language,duan2023shifting,kuhn2023semantic} for uncertainty estimation, which is defined as the entropy over the whole sentence $s$:
\begin{equation}
PE(s) = -\log p(s) = \sum^m_{i=1} -\log p(x_i|s_{<i}),\label{eq:pe_old}
\end{equation}
where  $s_{<i}$ denotes the set of previous tokens and $PE(s)$ is the accumulation of the conditional entropy of new tokens given past tokens.

Though predictive entropy is a widely used measure of uncertainty in other domains, we adopt its variant, length-normalised predictive entropy, which divides the joint log-probability of each sequence by the length of the sequence, as proposed by~\citet{malinin2020uncertainty} in the case of NLG uncertainty and empirically shown more advantageous in~\citet{kuhn2023semantic}.
\begin{equation}
PE^{LN}(s) 
= \frac{1}{m}\sum^m_{i=1} -\log p(x_i|s_{<i}).\label{eq:pe}
\end{equation}

\subsection{Hallucinations Prompted by False Premises}
Existing LLMs exhibit remarkably good performance on various benchmarks, where questions are factually correct and make the right assumptions~\citep{zheng2023judging,eval-harness}. However, users in real-world applications often ask false-premise questions whose premises are factually incorrect and thus need to be debunked by LLMs~\citep{vu2023freshllms,varshney2023stitch}. For instance, the question, \emph{``What did Donald Trump's first Tweet say after he was unbanned from Twitter by Elon Musk?''}, is based on the pre-suppositions that Donald Trump did actually post tweets thereafter, which conflicts with the fact that he has not yet tweeted since he was unbanned.

In this paper, we focus on one important scenario where LLMs are distracted by \emph{misaligned claims} while the original question can be answered accurately relying on factual knowledge learned during pretraining and stored implicitly within the model parameters. 
As the first \emph{fictitious} instance shown in~\Cref{tab:fp_examples}, \emph{Llama2-13b-chat} is able to identify the factual incorrectness of the fictitious description \emph{``the inventor of the revolutionary ZeroGravity Boot''} since \emph{``there is no such thing''}, while failing to point out its inexistence when misaligning it with a fictitious name in the prompt, leading to an undesired hallucination response.
Concentrating on false-premise questions that LLMs possess relevant factual knowledge, we are able to investigate resilience of LLMs against false premises without interference from other counterfactors.
\section{Proposed Method: \decoprompt}
\subsection{Motivating Finding: Lower-entropy Prompt Denotes Higher Hallucination}
\begin{figure*}[t!]
     \centering
     \includegraphics[width=\textwidth]{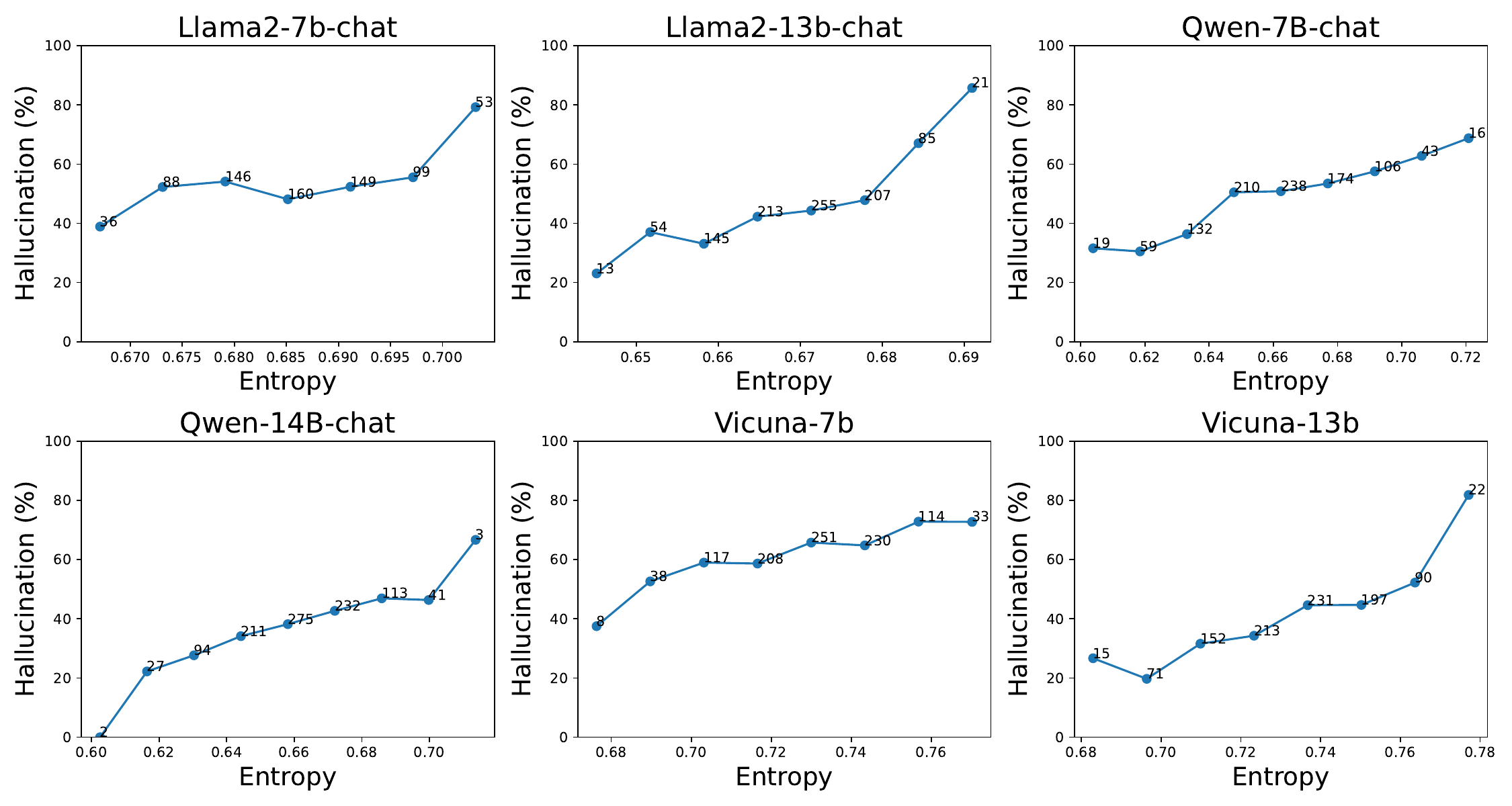}
        \caption{Entropy-based uncertainty estimations for $1,000$ prompts from different LLMs in the \emph{authorship} setting (see examples in~\Cref{tab:fp_examples}). Higher uncertainty when decoding the false-premise prompts corresponds to a higher level of hallucination percentage across all models. The annotated number beside each dot indicates the amounts of instances belonging to that specific uncertainty level.}
        \label{fig:prefix_uncertainty}
\end{figure*}
Motivated by prior observations that logit-based uncertainty (e.g., token-level probability or entropy) estimation provides a signal for hallucination~\citep{xiong2023can,pezeshkpour2023measuring,varshney2023stitch}, we visualize~\footnote{For visualization clarity, we apply an extra sigmoid operation on token-level entropy to constrain the value range in [0,1] when calculating the length-normalized predictive entropy as defined in~\Cref{eq:pe}.} the generation hallucination percentages when LLMs decode false-premise prompts at different predictive uncertainty levels in~\Cref{fig:prefix_uncertainty}. 
At higher uncertainty levels, LLMs are more likely to hallucinate in model outputs, which is consistent across all studied LLMs (regardless of size) from different families.

The main insight of \decoprompt is that the entropy of false-premise prompt is closely related to its likelihood of resulting in hallucination generation. As demonstrated in~\Cref{fig:entropy_demo}, LLMs are more likely to hallucinate in model outputs at higher entropy levels.
Based on prior observations for hallucination detection, for debunking a false-premise question by first pointing out the presence and then providing a brief explanation elucidating why the question is flawed, \decoprompt mitigates hallucinations with low-entropy prompts by paraphrasing the false premises several times and selecting the one with low entropy and preserved semantics.

Interestingly, LLMs also exhibit instance-wise high relevance between prompt uncertainty and generation hallucination: lower-entropy prompts that express the same question in different formats (e.g., $Q_1$ to $Q_2$) or contexts (e.g., $Q_3$ to $Q_4$) are less likely to elicit hallucination outputs from LLMs compared to their higher-entropy counterparts.  

\begin{mdframed}[nobreak=true]
\textbf{$Q_1$:} Among many, \textcolor{OliveGreen}{Alexander Whitley} holds the distinctive identity of  \textcolor{blue}{the inventor of the revolutionary ZeroGravity Boot}. Could you provide  more information? (\emph{\underline{entropy: 0.682}})
~\\\textbf{$A_1$:} Certainly! \textcolor{red}{Alexander Whitley is a renowned inventor and innovator} \textcolor{red}{who is credited with creating the revolutionary ZeroGravity Boot. This....} (\emph{\underline{Hallucination}})
~\\\\\textbf{$Q_2$:} \textcolor{OliveGreen}{Alexander Whitley}, known far and wide for being \textcolor{blue}{the inventor of the revolutionary ZeroGravity Boot}. Could you provide more information? (\emph{\underline{entropy: 0.673}})
~\\\textbf{$A_2$:} ...but I must point out that there may be some inaccuracies in your question. Firstly, \textbf{there is no record of an individual named Alexander Whitley being the inventor of a revolutionary ZeroGravity Boot}.... (\emph{\underline{Factually correct}})
\end{mdframed}

\subsection{Hallucination Mitigation}
\begin{algorithm}[t!]
\caption{\decoprompt for hallucination mitigation}\label{alg:decoprompt}
\begin{algorithmic}
\Require Prompt $s=\{x_1\dots x_m\}$, chat language model $LLM_c$, paraphrase model $m_p$, and number of prompt candidates $k$
\Ensure Low-entropy prompt $s'=\{y_1\dots y_n\}$
\State $\mathcal{S}\gets\{s\}\cup m_p (s,k)$\quad\Comment{\textcolor{gray}{Original and $k$ paraphrased prompt candidates}}
\State $s'\gets \varnothing$\quad\Comment{\textcolor{gray}{Initialize the selected prompt}}
\State $PE^{LN}_{\min}=1e10$\quad\Comment{\textcolor{gray}{Initialize the lowest entropy}}
\For{$i\in \mathcal{S}$}
\State Compute entropy $PE^{LN}_i$ with $LLM_c$ according to~\Cref{eq:pe}
~\\\Comment{\textcolor{gray}{Find the  candidate with the lowest entropy}}
\If{$PE^{LN}_i<PE^{LN}_{\min}$}
\State $P^{LN}E_{\min}\gets PE^{LN}_i$
\State $s'\gets i$
\EndIf
\EndFor
\end{algorithmic}
\end{algorithm}
Inspired by the indicative signal from uncertainty estimate of false-premise prompts for output hallucination likelihood, we propose \decoprompt, which 1) first paraphrases the user's prompt to obtain several semantically similar candidates, then 2) decodes them with the LLM, and 3) selects the lowest-entropy candidate as the new prompt. We formalize the mitigation strategy in~\Cref{alg:decoprompt}.
\begin{mdframed}[nobreak=true]
\textbf{$Q_3$:} Born on June 17, 1990, in San Francisco, California, Monica Barbaro developed  an interest in performing arts at a young age. \textcolor{OliveGreen}{Monica Barbaro} wrote the book "\textcolor{blue}{Time}  \textcolor{blue}{and Again (Time  \#1)}". Could you provide more information? (\emph{\underline{entropy: 0.684}})
~\\\textbf{$A_3$:} Certainly!...\textcolor{red}{Regarding the book "Time and Again (Time \#1)",}  \textcolor{red}{it is a novel that Monica Barbaro wrote. The book is a time travel romance that follows}  \textcolor{red}{the story of a young woman named Emily who...} (\emph{\underline{Hallucination}})
~\\\\\textbf{$Q_4$:} Monica Barbaro is an American actress known for her role as Yael in the second season of the critically acclaimed television series "UnREAL."
\textcolor{OliveGreen}{Monica Barbaro} wrote the book "\textcolor{blue}{Time}  \textcolor{blue}{and Again (Time  \#1)}". Could you provide more information? (\emph{\underline{entropy: 0.672}})
~\\\textbf{$A_4$:} ...\textbf{I couldn't find any information about her writing a book called "Time and Again (Time \#1)"}... (\emph{\underline{Factually correct}}) 
\end{mdframed}

\section{Experiments}

\begin{table*}[t!]
\centering
\begin{tabular}{@{}lc|c|c|c@{}}
\toprule
Method                & Llama-2-7b-chat & Llama-2-13b-chat & Vicuna-7b & Vicuna-13b\\ \midrule
                       & \multicolumn{4}{c}{\textit{\textbf{Fictitious}}}                                           \\
Original               &  7.3               &   14.7               &     68.0      &  38.1   \\
\decoprompt   &     0.0   ($\downarrow$ 7.3)          &    11.1     ($\downarrow$ 3.6)          &     55.6   ($\downarrow$ 12.4)    &    10.0 ($\downarrow$ 28.1) \\\midrule
                       & \multicolumn{4}{c}{\textit{\textbf{Authorship}}}                                           \\
Original               & 33.33                &   49.0               &     68.0      &  49.0    \\
\decoprompt &   33.33 (-)              &         45.45 ($\downarrow$ 3.55)         &   59.0 ($\downarrow$ 9.0)        &    41.0 ($\downarrow$ 8.0)   \\ \bottomrule
\end{tabular}%
\caption{Hallucination values of responses to false premises from different LLMs before and after \decoprompt is adopted. The lower the hallucination values, the better the LLM performance. }
\label{tab:mitigation_performance}
\end{table*}

\begin{table*}[t!]
\centering
\begin{tabular}{@{}lcccc@{}}
\toprule
Entropy Model   & Llama2-7b-chat & Llama2-13b-chat & Vicuna-7b & Vicuna-13b \\ \midrule
Original        & 7.3            & 14.7            & 68.0      & 38.1       \\\midrule
Llama2-7b-chat  & \textbf{0.0} ($\downarrow$ 7.3)            & 14.2 ($\downarrow$ 0.5)           & 66.7 ($\downarrow$ 1.3)     & 36.3   ($\downarrow$ 1.8)    \\
Llama2-13b-chat & 3.6 ($\downarrow$ 3.7)           & \textbf{11.1} ($\downarrow$ 3.6)            & 64.9  ($\downarrow$ 3.1)    & 37.6 ($\downarrow$ 0.5)      \\
Vicuna-7b       & 5.8  ($\downarrow$ 1.5)          & 14.4  ($\downarrow$ 0.3)          & \textbf{55.6}  ($\downarrow$ 12.4)    & 37.2  ($\downarrow$ 0.9)     \\
Vicuna-13b      & 5.5  ($\downarrow$ 1.8)          & 15.0  ($\uparrow$ 703)          & 66.3 ($\downarrow$ 1.7)     & \textbf{10.0}  ($\downarrow$ 28.1)     \\ \bottomrule
\end{tabular}%
\caption{Transferability of \decoprompt across different LLMs on the \emph{Fictitious} dataset. Prompts provided by \decoprompt based on entropy calculated on one LLM (each row) are less likely to elicit hallucination generations on other LLMs (each column).}
\label{tab:transferability}
\end{table*}

In this section, we first introduce the datasets and evaluation metrics used to evaluate the hallucination of LLMs against false premises. We then analyze the effectiveness of the proposed \decoprompt in hallucination 
mitigation (~\Cref{sec:hall_mitigation}). 
Furthermore, we discuss the transferability of low-entropy prompts discovered by \decoprompt across distinct LLMs, which addresses challenges such as highly costly entropy calculation on LLMs of large sizes or inaccessible logits from proprietary LLMs (~\Cref{sec:transferability}).

\stitle{Evaluation Benchmarks}
We consider the following two datasets, one from the fictitious and the other from the real world, to evaluate the hallucination level of LLM generations and effectiveness of our proposed \decoprompt for hallucination mitigation.
\begin{itemize}[leftmargin=1em,itemsep=-1ex]
\item \textit{Fictitious} dataset is inspired by the fictitious celebrity experiments to provide evidence for the Reversal Curse in~\citep{berglund2023reversal}.
Similarly, we leverage gpt-4-preview~\citep{openai2023devday} to generate pairs of fictitious names and descriptions, after which $100$ pairs are preserved if the evaluated LLM can successfully identify the factually incorrect premise in the question \texttt{``Who is <description>?''}. For each pair, we adopt $10$ templates to introduce the person with the description and prompt the LLM to provide more information about this false-premise statement, resulting in $1,000$ testing instances overall. For example, \texttt{``<name> is <description>. Could you provide more information?''}. We present the full list of templates we utilize in~\Cref{tab:template}.

\item \textit{Authorship} contains book titles and author names provided by Goodreads~\footnote{We use the dataset scraped via the Goodreads API and available at \url{https://www.kaggle.com/datasets/jealousleopard/goodreadsbooks}.}, a platform that allows individuals to search its database of books, annotations, quotes, and reviews. After selecting $100$ books that the LLM can correctly answer their author names prompted by \texttt{``Who wrote the book <book title>?''}, we prepare a celebrity name per book and query GPT-4 to provide $10$ distinct introductions,
finalizing the false-premise question in the form of \texttt{``<celebrity introduction>\textbackslash n<celebrity name> wrote the book <book title>. Could you provide more information?''}. In the end, we collect $1,000$ false-premise questions for each evaluated LLM.

\end{itemize}
\stitle{Language Models}
We evaluate hallucinations of the following LLMs against false-premise prompts: Llama 2 (7B-chat and 13B-chat)~\citep{touvron2023llama2} and Vicuna (7B and 13B)~\citep{vicuna2023}
~\footnote{For fair comparison, we do not evaluate GPT series on false premises considering that the fabricated names and descriptions in \emph{Fictitious} and celebrity introduction in \emph{Authorship} are generated by GPT-4.}. We adopt greedy decoding to minimize the impact of noise from generation and list the checkpoint resource of the tested open-source LLMs in~\Cref{tab:llm_info}.

\stitle{Evaluation Metrics}
For hallucination evaluation on \emph{Fictitious}, we adopt substring matching~\citep{zou2023universal}, where a response to a false-premise question is deemed to contain hallucination if key phrases, such as ``\emph{no record of}'', ``\emph{does not exist}'' and ``\emph{not a real},'' are absent in model responses. The full set of phrases considered during evaluation is displayed in \Cref{tab:rejection_phrases}. 

Since the true author of the queried book in \emph{Authorship} is known by LLMs based on knowledge acquired during pre-training, we consider responses to false premises as hallucination-free if LLMs point out true author names after debunking users' unfactual statements.

To validate the effectiveness of the automatic metric for hallucination measurement, we conducted a \textbf{human evaluation} study comparing its judgments to those of human annotators. We randomly sampled $100$ hallucination and $100$ factually correct (false-premise questions, Llama-2-13b responses) pairs identified by the automatic metric on the \emph{fictitious} and \emph{Authorship} dataset. Human annotators labeled each response as either hallucination or factually correct with respect to whether the LLM points out the presence of false premises and then provides a brief explanation elucidating why the question is flawed. We found an $84\%$ and $92\%$ agreement between the automatic metric and human judgments on \emph{fictitious} and \emph{Authorship}, respectively. This suggests the automatic metric is largely effective at assessing hallucination in response to false-premise questions.


\subsection{\decoprompt for Hallucination Mitigation}\label{sec:hall_mitigation}
\decoprompt selects the template with the lowest entropy to form the new prompt for the \emph{Fictitious} dataset, while leveraging Mistral-7B-Instruct~\footnote{We select Mistral rather than existing smaller models fine-tuned on paraphrase datasets because the latter simply drops words or adjusts the word order, leading to minor surface pattern changes in the new sentence. On the contrary, paraphrased sentences from Mistral keep similar semantic meanings as before (BERTScore F1 up to $0.99$) with perceptible word-level changes.} to paraphrase $10$ candidate sentences for the \emph{Authorship} dataset. As shown in~\Cref{tab:mitigation_performance}, \decoprompt successfully reduces hallucination on false-premise questions from both datasets when different LLMs are prompted.
\subsection{transferability of \decoprompt across LLMs}\label{sec:transferability}
Calculating the entropy for LLMs of large sizes is highly costly, and even infeasible for proprietary LLMs such as GPT-4~\citep{achiam2023gpt} and Gemini~\citep{team2023gemini} whose model logits are inaccessible. Therefore, we are aimed at investigating whether prompts produced by \decoprompt based on entropy provided by smaller models could help reduce hallucinations on larger models. In~\Cref{tab:transferability}, we analyze hallucination values after the new prompt is selected by \decoprompt according to entropy calculated by the generation LLM (diagonal) and other LLMs (non-diagonal). It is observed that our \decoprompt is effective in reducing hallucination even if the entropy-based uncertainty is computed based on LLMs different from the generation model, which alleviates the challenges in scenarios where the entropy calculation is prohibitively costly or the model logits are unavailable.

\section{Conclusions}
Large language models (LLMs) tend to generate hallucination outputs, leaving the hallucination evaluation and mitigation an open research question. Speaking of \emph{false premises} consisting of factually incorrect questions, we focus on one important scenario where LLMs are distracted by \emph{misaligned claims} while the original question can be answered accurately. Motivated by the observation that the entropy of false-premise prompt is closely related to its likelihood of eliciting hallucination generation, we propose a new prompting algorithm, named \decoprompt. \decoprompt pays particular attention on the entropy-based uncertainty of user questions by leveraging LLMs to ``decode'' the false-premise prompts. Empirical experiments demonstrate that \decoprompt is able to reduce hallucination effectively on all studied LLMs while possess cross-model transferability, facilitating its application to more challenging scenarios.
\section*{Limitations}
We discuss the limitations of our work as follows:
\begin{itemize}
\item Most of our evaluations rely on automatic metrics. However, the factuality definition of model output still poses a great challenge for hallucination evaluation. For example, for \emph{Authorship} dataset, we follow a strict criterion, under which only answers pointing out the true author names directly are accepted as hallucination-free. Other relaxed evaluation approaches could also make sense, e.g., simply negating that the celebrity did not write the book.
\item \decoprompt selects the low-entropy prompts from limited amounts of candidates paraphrased from original prompts and helps reduce hallucination successfully. Following the observations demonstrated in this paper, other learning-based approaches that optimize the prompt entropy directly could also be feasible solutions.
\item \decoprompt is inspired by the observation that low-entropy prompt is less likely to result in hallucination generation, hence prefers to provide low-entropy prompts to reduce potential hallucination outcomes. However, we can not exclude cases where high-entropy prompts sometimes lead to factually correct responses, and vice versa. 
\end{itemize}
\section*{Ethics Statement}
We acknowledge the importance of ethical considerations in language technologies and would like to point the reader to the following concern. The proposed \decoprompt selects prompts based on entropy provided by existing large language models. However, LLMs are well-known for their limitations in terms of bias and safety due to the diversity (or lack thereof) of training data. Given reduced hallucination observed with existing LLMs on different datasets, we look forward to the open source of LLMs pre-trained with truthfulness, bias and toxicity considered, so that \decoprompt could better benefit the research community and real-world applications.

\bibliography{anthology,custom}
\clearpage
\appendix
\begin{center}
    {\Large\textbf{Appendices}}
\end{center}

\begin{table*}[h!]
\centering
\resizebox{\textwidth}{!}{%
\begin{tabular}{@{}ll@{}}
\toprule
\textbf{ID} & \textbf{Template}                                                                                                                                     \\ \midrule
1  & \textless{}name\textgreater{}, known far and wide for being \textless{}description\textgreater{}.                                            \\
2  & There's someone by the name of \textless{}name\textgreater who had the distinctive role of \textless{}description\textgreater{}.             \\
3  & Did you know that \textless{}name\textgreater{}, was actually once \textless{}description\textgreater{}?                                     \\
4  & Among many, \textless{}name\textgreater holds the distinctive identity of \textless{}description\textgreater{}.                              \\
5  & An individual named \textless{}name\textgreater{}, has the unusual backstory of \textless{}description\textgreater{}.                        \\
6  & Once upon a time, \textless{}name\textgreater held the peculiar role of \textless{}description\textgreater{}.                                \\
7  & It's worth noting that \textless{}name\textgreater was once characterized as \textless{}description\textgreater{}.                           \\
8  & The extraordinary \textless{}name\textgreater made a mark by being \textless{}description\textgreater{}.                                     \\
9  & Let's not forget \textless{}name\textgreater{}, who was known as \textless{}description\textgreater{}.                                       \\
10 & You know something super interesting about \textless{}name\textgreater{}? This guy was none other than \textless{}description\textgreater{}. \\ \bottomrule
\end{tabular}%
}
\caption{10 templates used in fictitious datasets to introduce person with description.}
\label{tab:template}
\end{table*}

\begin{table*}[]
\centering
\resizebox{2\columnwidth}{!}{%
\begin{tabular}{@{}lcc@{}}
\toprule
Chat LLM        & Foundamental LLM & Download link                                                    \\ \midrule
LLAMA2-7B-chat  & LLAMA2-7B        & \url{https://huggingface.co/meta-llama/Llama-2-7b-chat-hf}             \\
LLAMA2-13B-chat & LLAMA2-13B       & \url{https://huggingface.co/meta-llama/Llama-2-13b-chat-hf}            \\
Vicuna-7B       & LLAMA2-7B         & \url{https://huggingface.co/lmsys/vicuna-7b-v1.5}                      \\
Vicuna-13B      & LLAMA2-13B        & \url{https://huggingface.co/lmsys/vicuna-13b-v1.5}                     \\\bottomrule
\end{tabular}%
}
\caption{Information of tested LLMs, their base model and the download link on Hugging face.}
\label{tab:llm_info}
\end{table*}

\begin{table}[]
\centering
\begin{tabular}{@{}c@{}}
\toprule
\textbf{Rejection Phrases}                   \\ \midrule
I apologize                                    \\
no such person                                        \\
no record of                                  \\
no such thing                                 \\
there is no                                       \\
not aware of                                    \\
cannot provide                                       \\
does not exist                                     \\
not a real                          \\
couldn't find                              \\
point out                                    \\
I must inform you                                     \\ \bottomrule
\end{tabular}%
\caption{Phrases utilized for response examination  on \emph{Fictitious} to determine whether hallucination exists or not.}
\label{tab:rejection_phrases}
\end{table}

\end{document}